\documentclass[11pt]{article}
\usepackage[preprint]{acl}
\usepackage{times}
\usepackage{latexsym}
\usepackage{enumitem}

\usepackage[no-math]{fontspec}
\usepackage{xeCJK}
\defaultCJKfontfeatures{}
\setCJKsansfont{FandolHei-Regular.otf}[BoldFont=FandolHei-Bold.otf]
\setCJKmonofont{FandolFang-Regular.otf}

\usepackage{amsmath,amsfonts,bm}
\renewcommand{\eqref}[1]{\textup{(\ref{#1})}}
\usepackage{graphicx}
\usepackage{booktabs}
\usepackage{multirow}
\usepackage{tabularx}
\usepackage{url}
\renewcommand{\abstractname}{Abstract}

\title{Learning to Reason Efficiently with Interventional Step Credit}
\author{
    Guanda Li\textsuperscript{1}\hspace{0.4em}%
    Xuefeng Bai\textsuperscript{1}\thanks{Corresponding author.}\hspace{0.4em}%
    Yulong Chen\textsuperscript{2,3}\hspace{0.4em}%
    Zhiyong Wang\textsuperscript{1}\hspace{0.4em}%
    Kehai Chen\textsuperscript{1}\hspace{0.4em}%
    Min Zhang\textsuperscript{1} \\
    \textsuperscript{1}Harbin Institute of Technology, Shenzhen \\
    \textsuperscript{2}University of Aberdeen \\
    \textsuperscript{3}University of Cambridge
}

\begin{document}
\maketitle

\begin{abstract}
Large reasoning models often produce lengthy chain-of-thought (CoT) traces mixing useful intermediate steps with unnecessary ones.
Recent methods explore step-level signals such as attention mass and step entropy to identify and suppress unnecessary steps. 
Despite their effectiveness, such signals reflect the model's behavior at a step rather than that step's contribution to the answer. 
To address this, we propose the interventional step credit (ISC), which takes a causal-intervention view and assesses step importance to the reasoning through targeted parameter-level interventions.
Building on ISC, we further propose IGPO, which uses ISC to guide fine-grained advantage rescaling in reinforcement learning, reducing redundant steps while preserving the important ones. 
Experiments across two model scales and nine benchmarks show that IGPO achieves up to 62\% average token reduction and better overall performance among compared methods.
\end{abstract}

\section{Introduction}
\label{sec:introduction}
In recent years, large reasoning models have achieved strong performance on complex reasoning tasks by generating long chains of thought (CoT) that verify intermediate results, reflect on earlier steps, and explore alternative solutions~\citep{wei2022chain,guo2025deepseekr1}.
However, such traces often exhibit ``overthinking'': they contain redundant computation in the form of repetitive derivations and unnecessary continuation, much of which substantially inflates inference cost while leaving the answer essentially unchanged~\citep{chen2025overthinking,jiang2026decs}.

A line of work seeks to suppress overthinking through reasoning budget control or length-aware reinforcement learning, encouraging models to answer both correctly and concisely~\citep{aggarwal2025l1,arora2025training,luo2025o1pruner,hou2025thinkprune,liu2025laser}.
However, such trajectory-level control makes limited distinctions between necessary and redundant steps, potentially suppressing both and compromising accuracy~\citep{nie2026attnpo}. 
Recent work therefore introduces finer-grained supervision. 
Early-exit methods identify a stopping point through prefix sufficiency or the first occurrence of a correct answer~\citep{dai2025sgrpo,tan2025depo,yue2026vsrm}, but leave redundancy within the retained prefix unresolved. 
Recent advances in step-level supervision score individual steps using attention mass~\citep{nie2026attnpo}, generation entropy~\citep{li2026stepentropy} and mean token log-probability~\cite{wang2026sas}. 
These signals characterize the model's behavior at a step rather than its dependence on that step for answer prediction: a low score does not reliably establish that suppressing the step will \mbox{preserve the answer}.\looseness=1

\begin{figure}[!t]
    \centering
    \includegraphics[width=\columnwidth]{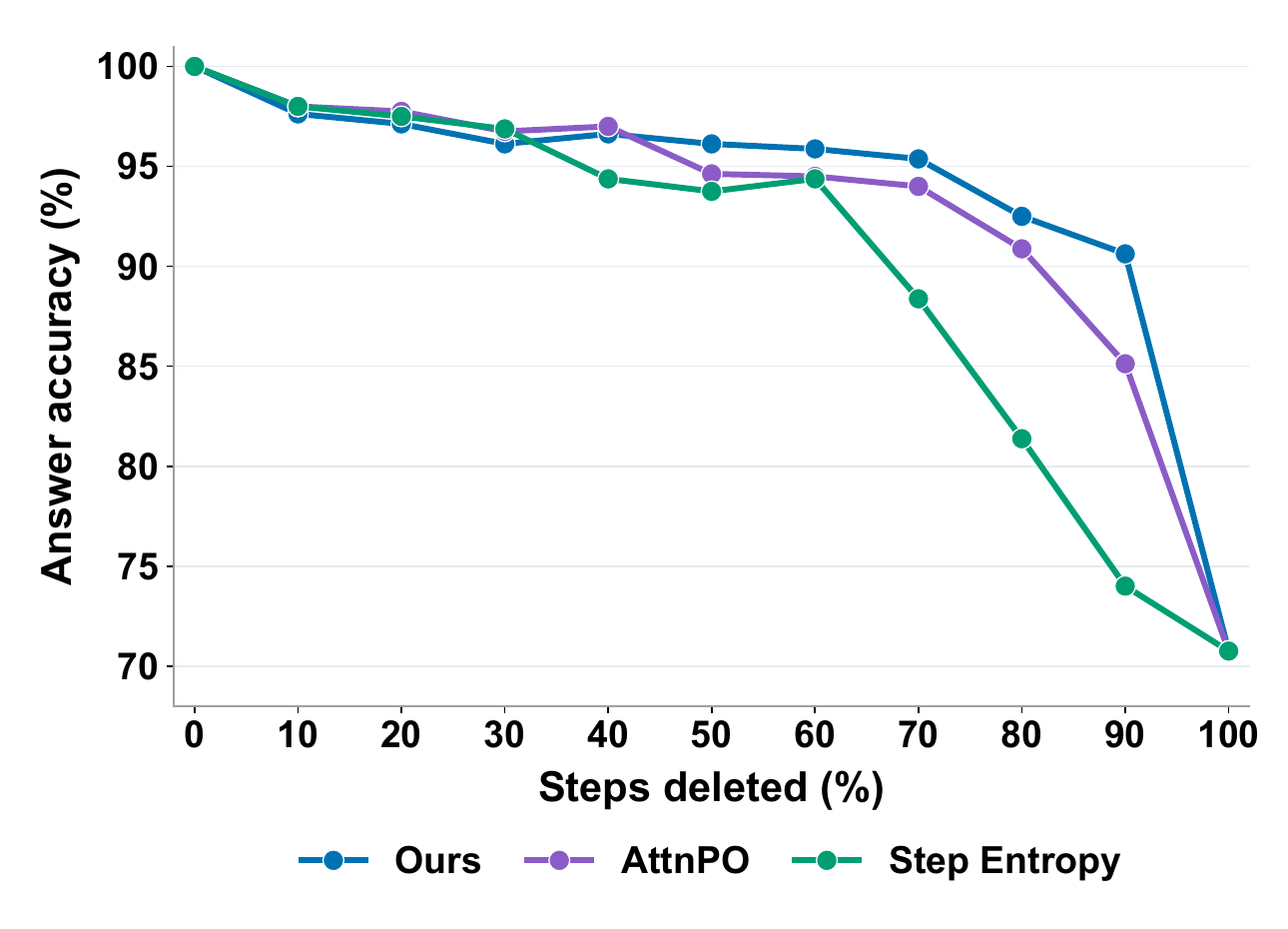}
    \caption{Answer accuracy on DS-7B under post-hoc CoT pruning. Deleting low-scoring steps identified by ISC better preserves accuracy than compared methods.}
    \label{fig:step-deletion}
\end{figure}

To address this, we propose interventional step credit (ISC), 
which takes a causal perspective on step importance through targeted parameter interventions.
Simply applying text-level step perturbations can obscure this influence, as the model may compensate for them when regenerating subsequent reasoning~\citep{tutek2025fur}.
We therefore probe this dependence through step-targeted parameter interventions, without regenerating subsequent reasoning.
Specifically, we use preference-based unlearning~\citep{zhang2024npo} to suppress the target step's generation under its preceding prefix.
With the prefix and the base model's candidate answer fixed, ISC measures the magnitude of the resulting change in answer confidence.
As shown in Figure~\ref{fig:step-deletion}, ISC generally achieves a more favorable accuracy–pruning frontier than attention mass~\citep{nie2026attnpo} or step entropy~\citep{li2026stepentropy} in post-hoc CoT pruning, particularly at high step-deletion ratios.

Building on ISC, we propose IGPO, which distills intervention-derived step-credit labels into an online step-level predictor and uses the predicted credit to guide stepwise advantage rescaling. 
For correct trajectories, IGPO attenuates the positive advantage of low-ISC steps and shields high-ISC steps from negative advantage.
This selective rescaling discourages reinforcing redundant reasoning while protecting important steps, enabling the model to produce shorter reasoning traces while maintaining \mbox{strong task performance}.\looseness=1

We evaluate the proposed method on two representative LLMs across nine benchmarks. 
Experimental results show that IGPO reduces average response length by up to 62.0\% while improving macro-average accuracy by up to 8.4~pp over base models.
IGPO also achieves the highest macro-average accuracy and efficiency score among the compared methods at two model scales.
Further analysis shows that IGPO generates more concise CoT with more reliable computations and logical deductions than \mbox{compared methods}.\looseness=1

Our main contributions are as follows:
\begin{itemize}[itemsep=2pt,topsep=0pt,parsep=0pt,leftmargin=11pt]
    \item We introduce ISC, a causal formulation of step importance grounded in targeted parameter-level interventions.
    \item We propose IGPO, which employs ISC to selectively rescale stepwise advantages, targeting redundant steps while protecting important steps.
    \item Experiments show that ISC improves the accuracy--length trade-off in post-hoc CoT pruning, while IGPO reduces reasoning costs and improves task performance.
\end{itemize}

\section{Related Work}
\label{sec:related}

\paragraph{Efficient Reasoning.}
Existing methods are broadly categorized as training-free and training-based. Training-free methods reduce inference costs through budget control, early termination, or CoT pruning~\cite{han2025tale,yang2025deer,choi2025thinkclearly,huang2026sat}. 
Training-based methods use trajectory-level policy optimization~\cite{arora2025training,aggarwal2025l1,liu2025laser,wu2025lapo,huang2026sage} or finer-grained policy optimization~\cite{dai2025sgrpo,li2026swap,nie2026attnpo,xu2026mpo}. 
Our work uses intervention-derived step credit to guide advantage rescaling and selectively discourage redundant reasoning.

\paragraph{Step Importance Assessment.}
Prior work assesses reasoning-step importance through answer position, prefix-conditioned performance, or model-internal signals.
Answer-position methods use the first correct answer as a stopping boundary, leaving redundancy within the retained prefix unresolved~\citep{cheng2025lcr1,jiang2026decs,tan2025depo}.
Prefix-based methods evaluate progress toward the ground-truth answer~\citep{kazemnejad2025vineppo,setlur2025rewarding,dai2025sgrpo,yue2026vsrm,li2026swap,wu2026spae}, but may undervalue preparatory steps whose contributions have not yet improved answer predictions.
Recent advances introduce step scores based on attention, generation entropy, and token-level confidence~\citep{nie2026attnpo,li2026stepentropy,wang2026sas}.
These signals characterize model behavior rather than directly measuring its dependence on a step for answer prediction.
Most related to us, SWAP~\citep{li2026swap} and SPAE~\citep{wu2026spae} use answer predictions to assess each step's contribution to reasoning progress.
ISC instead assesses how answer prediction depends on the target step through parameter intervention, while holding the prefix and the answer fixed.
This directly captures the target step's causal effect on the reasoning state, revealing the true value of preparatory steps.

\paragraph{Interventions on Reasoning Steps.}
Prior work primarily uses text-level interventions, either perturbing steps and regenerating the continuation~\citep{lanham2023faithfulness,bogdan2025thoughtanchors,yu2025causalsufficiency,ruan2025cft}, or deleting text units and measuring prediction changes conditioned on the remaining CoT, as in SCPO~\citep{he2025smartthinker}, AttriCoT~\citep{wei2026attricot}, and LIMOPro~\citep{xiao2025limopro}.
However, regenerated continuations may compensate for perturbations through correction or alternative reasoning, while retained downstream steps may still carry information supplied by the deleted step.
\citet{li2026latentdynamics} intervene on steps by modifying hidden states.
FUR~\citep{tutek2025fur} uses parameter interventions to assess the parametric faithfulness of reasoning steps through prediction changes in a no-CoT setting.
However, a mathematical reasoning step's role often depends on previously accumulated information, which no-CoT evaluation may not fully capture.
We follow the key idea of FUR but evaluate answer confidence under the fixed prefix preceding the target step, without regenerating subsequent reasoning.
ISC thus assesses prefix-conditioned step influence rather than no-CoT parametric faithfulness.

\section{Interventional Step Credit}
\label{sec:method}
\label{sec:isc}

\begin{figure*}[!t]
    \centering
    \includegraphics[width=\textwidth]{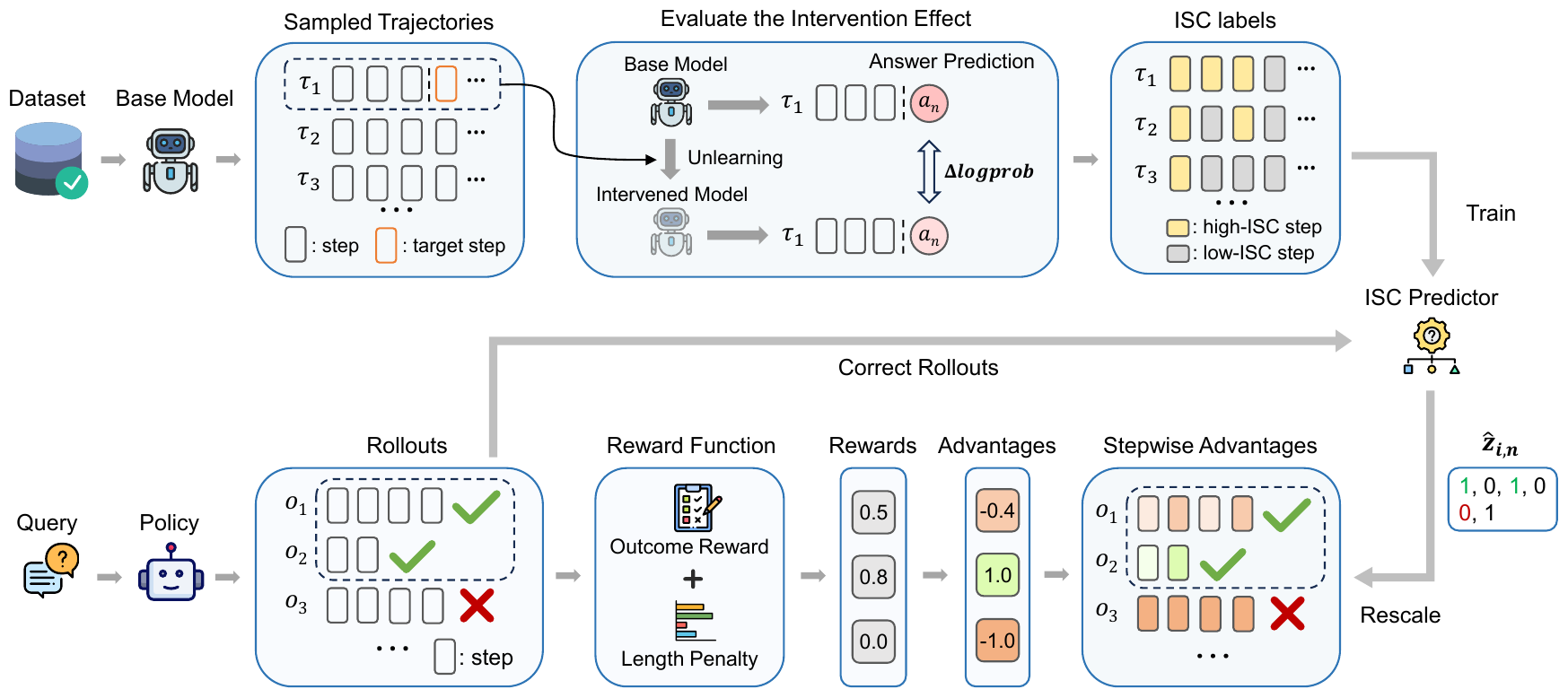}
    \caption{The overall framework of ISC and IGPO}
    \label{fig:main-method}
\end{figure*}

We introduce interventional step credit (ISC) to assess the influence of a reasoning step through targeted parameter interventions.
We formulate each reasoning step's causal contribution to the answer (\S\ref{sec:causal-formulation-of-step-importance}), then estimate ISC from intervention-induced changes in answer confidence (\S\ref{sec:ISC-estimation}).

\subsection{Causal Formulation of Step Importance}
\label{sec:causal-formulation-of-step-importance}

Given a question $q$, the model generates a response $o=(\tau,\mathcal{F})$, where $\tau=(s_1,\ldots,s_N)$ denotes a reasoning trajectory of $N$ steps and $\mathcal{F}$ denotes the final solution.
For the $n$-th step $s_n$, we define its preceding reasoning prefix as $h_n=(q,s_1,\ldots,s_{n-1})$.
We use $|o|$ to denote the number of tokens in the complete response.
Given $h_n$, we examine the causal dependence of the model's answer prediction on the computation underlying step $s_n$.
Specifically, we assume that after processing the reasoning prefix $h_n$,
the model has formed a latent prediction of the next reasoning step $\tilde{s}_n$
and an anticipated subsequent reasoning trajectory $\tilde{s}_{>n}$.
The actual step $s_n$ is the verbalized realization of $\tilde{s}_n$, 
while the anticipated subsequent reasoning $\tilde{s}_{>n}$ supports the model's answer prediction $\hat{a}_n$ under prefix $h_n$.
We use two idealized causal structures to describe how $\tilde{s}_n$ influences answer prediction through subsequent reasoning.

In the first structure, $\tilde{s}_n$ is verbalized as $s_n$ and further shapes the model's anticipated subsequent reasoning $\tilde{s}_{>n}$ and answer prediction $\hat{a}_n$:
\begin{equation}
    \begin{array}{ccccccc}
        h_n & \longrightarrow & \tilde{s}_n & \longrightarrow
        & \tilde{s}_{>n} & \longrightarrow & \hat{a}_n \\
        & & \downarrow & & & & \\
        & & s_n & & & &
    \end{array}
    \label{eq:important-step-causal-chain}
\end{equation}

In the second structure, $\tilde{s}_n$ is still verbalized as $s_n$, while the anticipated subsequent reasoning $\tilde{s}_{>n}$ and answer prediction $\hat{a}_n$ are driven primarily by the information in $h_n$:
\begin{equation}
    \begin{array}{ccccc}
        h_n & \longrightarrow & \tilde{s}_{>n} & \longrightarrow & \hat{a}_n \\
        \downarrow & & & & \\
        \tilde{s}_n & \longrightarrow & s_n & &
    \end{array}
    \label{eq:redundant-step-causal-chain}
\end{equation}

From this causal perspective, we characterize step importance by whether the computation underlying a step mediates the model's answer prediction.
In the next subsection, we measure this influence through intervention and quantify it as \textbf{Interventional Step Credit (ISC)}.
Intervening on the computation underlying an important step should induce a substantial change in answer prediction, yielding high ISC.

\subsection{ISC Estimation}
\label{sec:ISC-estimation}

\paragraph{Step-Targeted Parameter Intervention.}
We implement step-targeted parameter interventions by unlearning the entire step $s_n$ conditioned on $h_n$, using the NPO+KL algorithm employed in FUR~\citep{tutek2025fur}.
The NPO term~\citep{zhang2024npo} reduces the probability of generating $s_n$, thereby approximately suppressing its underlying latent computation $\tilde{s}_n$.
The KL term regularizes predictions toward those of the base model on a retain set $\mathcal{D}_{\mathrm{RT}}$ drawn from other reasoning trajectories.
We minimize the joint objective:
\begin{equation}
    \mathcal{L}_{\mathrm{U}}^{(n)}(\theta)
    =\mathcal{L}_{\mathrm{NPO}}^{(n)}(\theta)
    +\lambda_{\mathrm{KL}}\mathcal{L}_{\mathrm{KL}}(\theta),
    \label{eq:step-unlearning-objective}
\end{equation}
where $\lambda_{\mathrm{KL}}$ controls the strength of retain regularization.
Both loss terms are detailed in Appendix~\ref{app:annotation-details}.
For each step, we independently update the base parameters $\theta_0$ to obtain $\theta^{(-n)}$.

\paragraph{Measuring the Intervention Effect.}
For a target step $s_n$, we prompt the base model $\pi_{\theta_0}$ to answer directly given the reasoning prefix $h_n$, yielding $a_n$.
Keeping both $h_n$ and $a_n$ fixed,
we then evaluate the confidence assigned to $a_n$ by the base model $\pi_{\theta_0}$ and the intervened model $\pi_{\theta^{(-n)}}$.
We avoid regenerating subsequent reasoning to isolate the intervention effect from changes in reasoning paths and contexts.
Suppose that $a_n$ consists of $T_n$ tokens. We use its average log-probability as the model's confidence in the answer:
\begin{equation}
    \ell_{\theta}(a_n\!\mid\! h_n)
    =\frac{1}{T_n}\sum_{t=1}^{T_n}\!
    \log\pi_{\theta}(a_{n,t}\!\mid\! h_n,a_{n,<t}),
    \label{eq:direct-answer-confidence}
\end{equation}
where $a_{n,t}$ denotes the $t$-th token of $a_n$.
We define the ISC of $s_n$ as the magnitude of the intervention-induced change in answer confidence:
\begin{equation}
    c_n=\left|
    \ell_{\theta_0}(a_n\mid h_n)
    -\ell_{\theta^{(-n)}}(a_n\mid h_n)
    \right|.
    \label{eq:causal-step-credit}
\end{equation}

\section{IGPO: ISC-Guided Policy Optimization}
\label{sec:igpo}

IGPO turns ISC into step-level supervision for efficient reasoning.
We construct binary ISC labels and distill them into a step-level predictor (\S\ref{sec:ISC-distillation}).
The predicted credit guides stepwise advantage rescaling during reinforcement learning (\S\ref{sec:ISC-guided-policy-optimization}).

\subsection{ISC Labeling and Distillation}
\label{sec:ISC-distillation}

To avoid costly online parameter interventions, we distill offline ISC labels into a step-level predictor.

\paragraph{ISC Labeling.}
For $K$ annotation trajectories, let $N_i$ be the number of steps in trajectory $i$ and $c_{i,n}$ the ISC of its $n$-th step.
We pool ISC values from all steps across all trajectories into $\mathcal{C}=\{c_{i,n}\}$ and label the global top $\rho$ fraction as high-ISC:
\begin{equation}
    z_{i,n}
    =
    \mathbb{I}
    \left[
        c_{i,n}
        \ge
        Q_{1-\rho}(\mathcal{C})
    \right],
    \label{eq:ISC-label}
\end{equation}
where $\rho\in(0,1)$, $Q_{1-\rho}$ denotes the corresponding quantile, and $z_{i,n}=1$ indicates a high-ISC step.

\paragraph{ISC Prediction.}
A causal LM with a shared classification head predicts $p_{i,n}=g_{\phi}(h_{i,n},s_{i,n})$, the probability that a completed step $s_{i,n}$ is high-ISC given its prefix $h_{i,n}$.
We train the predictor with binary cross-entropy, averaging the loss over steps within each trajectory and then across trajectories (Appendix~\ref{app:predictor-details}).
During policy optimization, we use $\widehat{z}_{i,n}=\mathbb{I}[p_{i,n}\geq\eta]$ to identify high-ISC steps, where $\eta$ is the prediction threshold.

\subsection{ISC-Guided Policy Optimization}
\label{sec:ISC-guided-policy-optimization}

We use predicted ISC to guide length-aware reinforcement learning.
We adopt a trajectory-level reward that combines answer correctness with a relative length penalty
and perform stepwise advantage rescaling according to the predicted ISC.

\paragraph{Trajectory-Level Reward and Advantage Estimation.}
Given a question $q$, the old policy $\pi_{\psi_{\mathrm{old}}}$ samples a group of $G$ rollouts $\{o_i\}_{i=1}^{G}$.
Following TLMRE~\citep{arora2025training}, we define a trajectory-level reward that combines answer correctness with a relative length penalty:
\begin{equation}
    r_i
    =
    \mathbb{I}[o_i\text{ is correct}]
    \left(
        1-\alpha\,\sigma\!\left(f(o_i)\right)
    \right),
    \label{eq:trajectory-reward}
\end{equation}
\begin{equation}
    f(o_i)
    =
    \frac{|o_i|-\operatorname{mean}(q)}
    {\operatorname{std}(q)},
    \label{eq:normalized-length}
\end{equation}
where $|o_i|$ denotes the response length,
$\operatorname{mean}(q)$ and $\operatorname{std}(q)$ denote the mean and standard deviation of the lengths
of correct rollouts for question $q$, respectively,
$\sigma(\cdot)$ is the sigmoid function, and $\alpha$ controls the strength of the length penalty.
Based on these trajectory-level rewards, we estimate the advantage using RLOO:
\begin{equation}
    A_i
    =
    r_i
    -
    \frac{1}{G-1}
    \sum_{j\neq i} r_j.
    \label{eq:rloo-advantage}
\end{equation}
The trajectory-level advantage $A_i$ is shared across all reasoning steps.
To focus length optimization on low-ISC steps while protecting high-ISC steps, we introduce stepwise advantage rescaling.

\paragraph{ISC-Guided Advantage Rescaling.}
We apply stepwise advantage rescaling only to correct rollouts, using predicted ISC labels $\widehat{z}_{i,n}$.
Incorrect rollouts retain the original trajectory-level advantage $A_i$.
For positive advantages,
we attenuate the positive updates assigned to low-ISC steps to avoid reinforcing redundant reasoning.
Conversely, for negative advantages,
we suppress the negative updates on high-ISC steps to protect essential reasoning from excessive length penalties.
We set $\widetilde{A}_{i,n}=\gamma_{i,n}A_i$ with
\begin{equation}
    \gamma_{i,n}=
    \begin{cases}
        \delta, & A_i>0,\ \widehat{z}_{i,n}=0, \\
        0, & A_i<0,\ \widehat{z}_{i,n}=1, \\
        1, & \text{otherwise}.
    \end{cases}
    \label{eq:ISC-advantage-shaping}
\end{equation}
Here, $\delta \in [0,1]$ scales positive advantages for low-ISC steps.
Since $\gamma_{i,n}\geq0$, rescaling does not reverse their signs.
Each token in $s_{i,n}$ receives $\tilde{A}_{i,n}$, yielding token-level advantages $\tilde{A}_{i,t}$ for the clipped policy objective (Appendix~\ref{app:policy-details}).

\section{Experiments}
\label{sec:experiments}
\subsection{Experimental Setup}
\label{sec:experimental-setup}

\paragraph{Training.}
We train DeepSeek-R1-Distill-Qwen-1.5B (DS-1.5B) and DeepSeek-R1-Distill-Qwen-7B (DS-7B)~\citep{guo2025deepseekr1} with IGPO on the DeepScaleR dataset~\citep{luo2025deepscaler}
using the VERL framework~\citep{sheng2025hybridflow}.
We set $\alpha=0.15$ for both models (Eq.~\ref{eq:trajectory-reward}), with $\delta=0.7$ for DS-1.5B and $\delta = 0.5$ for DS-7B.

\paragraph{Evaluation.} We evaluate on six in-domain reasoning benchmarks: GSM8K~\citep{cobbe2021verifiers}, MATH500~\citep{lightman2024verify}, AMC23~\citep{mathai2025amc23}, OlympiadBench~\citep{he2024olympiadbench}, AIME2024, and AIME2025~\cite{maa2024aime,maa2025aime}.
In addition, we report performance on three OOD benchmarks: LiveCodeBench-v6~\citep{jain2025livecodebench}, GPQA-Diamond~\citep{rein2024gpqa}, and MMLU~\citep{hendrycks2020mmlu}, covering math, coding, and science tasks of varying difficulty.
To obtain more reliable estimates, we evaluate each problem 128 times on AIME2024 and AIME2025 and four times on other benchmarks, and report the average performance.
We report Pass@1 (Acc.), response length (Tok.), and compute the Accuracy--Efficiency Score (AES)~\citep{luo2025o1pruner} to summarize the accuracy–length trade-off.

\paragraph{Baselines.}
We compare against three categories of efficient reasoning methods:
(1) adaptive reasoning methods, including AdaptThink~\citep{zhang2025adaptthink}, AutoThink~\citep{tu2025autothink}, and DEER~\citep{yang2025deer}, included for reference;
(2) outcome-supervised RL methods, including TLMRE~\citep{arora2025training}, ThinkPrune~\citep{hou2025thinkprune}, DIET~\citep{chen2025diet}, ACPO~\citep{cheng2025acpo}, LASER~\citep{liu2025laser} (LASER-D and LASER-DE), SmartThinker~\citep{hu2026smartthinker}, and Step Entropy~\citep{li2026stepentropy};
and (3) process-supervised RL methods, including LC-R1~\citep{cheng2025lcr1}, VSRM-R++~\citep{yue2026vsrm}, S-GRPO~\citep{dai2025sgrpo}, DEPO~\citep{tan2025depo}, DECS~\citep{jiang2026decs}, SWAP~\citep{li2026swap}, SPAE~\citep{wu2026spae}, and AttnPO~\citep{nie2026attnpo}.

\paragraph{ISC Annotation and Distillation.}
For each model, we generate 1,000 correct trajectories on DeepScaleR for ISC annotation and construct binary labels using $\rho=0.4$ (Eq.~\ref{eq:ISC-label}).
We train an ISC predictor based on Qwen3-4B-Instruct-2507~\citep{qwen2025qwen3} on 900 trajectories and reserve 100 for validation.
Implementation details are provided in Appendices~\ref{app:details} and~\ref{app:experimental-settings}.

\subsection{ISC Evaluation}
\label{sec:ISC-analysis}
We evaluate ISC through post-hoc CoT pruning, answer-sensitivity analysis, and checks for non-target intervention effects.

\paragraph{Post-hoc CoT pruning.} We assess whether ISC identifies steps that can be removed while preserving answer accuracy. 
We sample 50 \textit{correct} trajectories generated by DS-7B on DeepScaleR-Preview. 
Following~\citealt{li2026stepentropy}, we remove increasing fractions of the lowest-scoring steps under ISC, AttnPO, and Step Entropy, and evaluate answer accuracy conditioned on the remaining CoT. 
We always mask the final-answer step to prevent direct answer reuse. 
Figure~\ref{fig:step-deletion} shows that ISC better preserves accuracy under aggressive pruning, supporting its usefulness for identifying removable reasoning steps.

\begin{figure}[!t]
    \centering
    \includegraphics[width=\columnwidth]{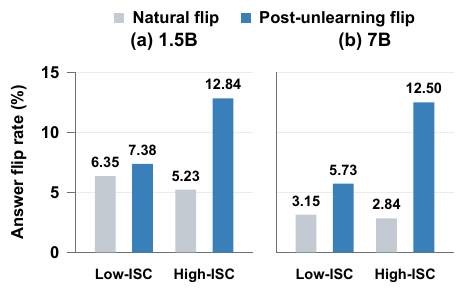}
    \caption{Natural and post-unlearning answer flip rates for low- and high-ISC steps on the 1.5B and 7B models.}
    \label{fig:isc-answer-changes}
\end{figure}

\paragraph{Discriminative Ability of ISC.} We randomly sample 100 annotated trajectories and select one high-ISC and one low-ISC step from each. For each step, we compare four base-model and four intervened-model completions against four base-model reference completions under the same prefix. 
We define the \textit{excess flip rate} as the post-unlearning flip rate minus the base model's natural flip rate.
As shown in Figure~\ref{fig:isc-answer-changes}, high-ISC interventions yield much larger excess flip rates than low-ISC interventions at both model scales, demonstrating ISC's ability to identify steps whose unlearning more strongly affects answer predictions.

\begin{figure}[!t]
    \centering
    \includegraphics[width=\columnwidth]{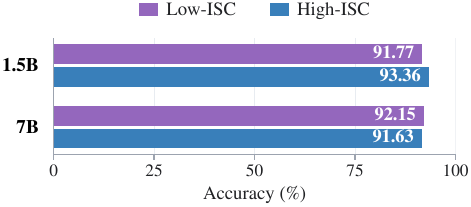}
    \caption{Accuracy on unrelated questions after unlearning low- and high-ISC steps in the 1.5/7B models.}
    \label{fig:unrelated-question-retention}
\end{figure}

\begin{figure}[t]
    \centering
    \includegraphics[width=\columnwidth]{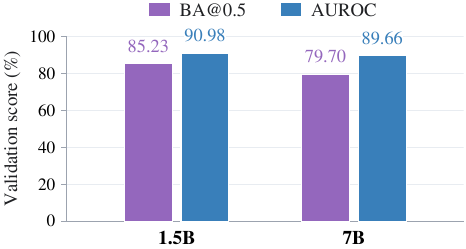}
    \caption{Predictor performance on the validation set for labels
    derived from the 1.5B/7B models. BA@0.5 denotes balanced accuracy
    at a threshold of 0.5.}
    \label{fig:credit-predictor-evaluation}
\end{figure}

\paragraph{Assessing Potential Confounds.}
To rule out potential confounding effects, we sample 25 unrelated problems answered correctly in all four sampled base-model responses. 
For each intervened model, we sample four responses per problem and average accuracy within each ISC group. Figure~\ref{fig:unrelated-question-retention} shows similar accuracy between high-ISC and low-ISC interventions at both model scales. 
These results suggest that the larger answer changes after high-ISC interventions are not simply due to greater overall capability degradation.

\paragraph{ISC Predictor Evaluation.} We train the ISC predictor for 10 epochs and select the checkpoint with the highest balanced accuracy on 100 held-out validation trajectories for policy optimization. 
Figure~\ref{fig:credit-predictor-evaluation} reports BA@0.5 and AUROC, showing discrimination between high- and low-ISC steps on labels derived from both model scales.

\begin{table*}[!t]
    \centering
    \begingroup
    \scriptsize
    \setlength{\tabcolsep}{1.4pt}
    \begin{tabular*}{\textwidth}{@{\extracolsep{\fill}}l*{6}{rrr@{\hspace{3pt}}}rrr@{}}
        \toprule
        \multirow{2}{*}[-2.5pt]{\textbf{Methods}} & \multicolumn{3}{c}{\textbf{GSM8K}} & \multicolumn{3}{c}{\textbf{MATH500}} & \multicolumn{3}{c}{\textbf{AMC2023}} & \multicolumn{3}{c}{\textbf{Olympiad}} & \multicolumn{3}{c}{\textbf{AIME2024}} & \multicolumn{3}{c}{\textbf{AIME2025}} & \multicolumn{3}{c}{\textbf{Macro Avg}} \\
        \cmidrule(l{1.4pt}r{3pt}){2-4} \cmidrule(l{0pt}r{3pt}){5-7} \cmidrule(l{0pt}r{3pt}){8-10} \cmidrule(l{0pt}r{3pt}){11-13} \cmidrule(l{0pt}r{3pt}){14-16} \cmidrule(l{0pt}r{3pt}){17-19} \cmidrule(l{0pt}r{0pt}){20-22}
        & Acc. & Tok. & AES & Acc. & Tok. & AES & Acc. & Tok. & AES & Acc. & Tok. & AES & Acc. & Tok. & AES & Acc. & Tok. & AES & Acc. & Tok. & AES \\
        \midrule
        \multicolumn{22}{c}{\textbf{1.5B Model}} \\
        \midrule
        DS-1.5B & 78.9 & 1018 & 0.00 & 81.7 & 4670 & 0.00 & 67.5 & 7356 & 0.00 & 48.6 & 9066 & 0.00 & 27.6 & 12327 & 0.00 & 22.9 & 12086 & 0.00 & 54.5 & 7754 & 0.00 \\
        $\heartsuit$ AutoThink & 83.0 & 568 & 0.60 & 84.0 & 2195 & 0.62 & 67.0 & 5059 & 0.28 & 51.6 & 5053 & 0.63 & 34.6 & 9514 & 0.99 & 21.8 & 7944 & 0.10 & 57.0 & 5056 & 0.54 \\
        $\heartsuit$ AdaptThink & 83.1 & \underline{480} & 0.69 & 82.0 & 1782 & 0.63 & 67.0 & 3510 & 0.49 & 50.8 & 3708 & 0.73 & 31.0 & 6670 & 0.83 & 23.0 & 7255 & 0.41 & 56.1 & 3901 & 0.63 \\
        $\heartsuit$ DEER$^{\ddagger}$ & 74.7 & 984 & -0.23 & 67.8 & 2497 & -0.38 & 60.0 & 5496 & -0.30 & 32.0 & 5960 & -1.37 & 23.3 & 9557 & -0.55 & 10.0 & 9281 & -2.58 & 44.6 & 5629 & -0.90 \\
        $\spadesuit$ ThinkPrune & 83.0 & 609 & 0.56 & 83.2 & 1938 & 0.64 & 73.2 & 3039 & 0.84 & 51.4 & 3535 & 0.78 & 27.1 & 5631 & 0.46 & 22.5 & 5167 & 0.49 & 56.7 & 3320 & 0.63 \\
        $\spadesuit$ DIET & -- & -- & -- & 83.0 & 3061 & 0.39 & 65.4 & 6425 & -0.03 & -- & -- & -- & 31.8 & 10578 & 0.60 & -- & -- & -- & -- & -- & -- \\
        $\spadesuit$ ACPO & 81.3 & 572 & 0.53 & 81.0 & 1679 & 0.60 & -- & -- & -- & -- & -- & -- & 30.0 & 6670 & 0.72 & -- & -- & -- & -- & -- & -- \\
        $\spadesuit$ Laser-D & 83.4 & 863 & 0.33 & 84.2 & 1872 & 0.69 & 75.3 & 2981 & 0.94 & 54.4 & 4700 & 0.84 & 34.2 & 5750 & 1.25 & 23.1 & 6928 & 0.45 & 59.1 & 3849 & 0.75 \\
        $\spadesuit$ Laser-DE & 80.4 & 820 & 0.25 & 83.5 & 1949 & 0.65 & 73.3 & 3080 & 0.84 & 54.4 & 5151 & 0.79 & 35.0 & 5789 & 1.34 & 24.2 & 7323 & 0.57 & 58.5 & 4019 & 0.74 \\
        $\spadesuit$ SmartThinker & 79.0 & 697 & 0.32 & 82.0 & 2516 & 0.47 & 71.3 & 4022 & 0.62 & 52.5 & 4982 & 0.69 & 26.7 & 8022 & 0.19 & 21.4 & 7369 & 0.05 & 55.5 & 4601 & 0.39 \\
        $\clubsuit$ LC-R1 & 82.7 & 841 & 0.32 & 82.5 & 2233 & 0.55 & 61.7 & 3947 & 0.03 & 48.1 & 4546 & 0.45 & 23.6 & 7122 & -0.30 & 21.2 & 6434 & 0.10 & 53.3 & 4187 & 0.19 \\
        $\clubsuit$ VSRM-R++ & -- & -- & -- & 81.7 & 2597 & 0.45 & 64.7 & 4119 & 0.23 & 54.8 & 4388 & 0.90 & 29.5 & 6958 & 0.64 & 22.9 & 6892 & 0.43 & -- & -- & -- \\
        $\clubsuit$ DEPO & -- & -- & -- & \textbf{87.2} & 2762 & 0.61 & 74.2 & 4388 & 0.70 & -- & -- & -- & 30.8 & 7732 & 0.72 & 24.8 & 7649 & 0.62 & -- & -- & -- \\
        $\clubsuit$ DECS & -- & -- & -- & 84.4 & 1817 & 0.71 & 75.4 & 2988 & 0.94 & 56.1 & 3396 & 1.09 & 31.2 & \underline{5550} & 0.94 & 23.8 & \underline{4965} & 0.71 & -- & -- & -- \\
        $\clubsuit$ SWAP & -- & -- & -- & \underline{86.5} & \underline{1597} & \underline{0.84} & 78.3 & \underline{2645} & \underline{1.12} & 45.1 & 3646 & 0.24 & 33.9 & \textbf{4915} & 1.29 & 23.3 & \textbf{4627} & 0.67 & -- & -- & -- \\
        $\spadesuit$ TLMRE & \underline{87.2} & 604 & 0.72 & 85.8 & 1915 & 0.74 & \underline{79.1} & 3349 & 1.06 & 56.9 & 4270 & 1.04 & \underline{37.3} & 7024 & 1.49 & \textbf{26.6} & 7117 & \underline{0.90} & \underline{62.2} & 4046 & 0.99 \\
        $\clubsuit$ AttnPO & 87.0 & \textbf{393} & \textbf{0.92} & 86.0 & \textbf{1318} & \textbf{0.88} & 77.8 & \textbf{2638} & 1.10 & \underline{57.1} & \textbf{3133} & \underline{1.18} & \textbf{37.7} & 5713 & \textbf{1.64} & \underline{25.4} & 5488 & 0.87 & 61.8 & \textbf{3114} & \underline{1.10} \\
        \midrule
        $\clubsuit$ \textbf{Ours} & \textbf{87.4} & 586 & \underline{0.75} & \underline{86.5} & 1662 & 0.82 & \textbf{81.5} & 2798 & \textbf{1.24} & \textbf{57.7} & \underline{3359} & \textbf{1.19} & \textbf{37.7} & 6027 & \underline{1.61} & \textbf{26.6} & 5481 & \textbf{1.03} & \textbf{62.9} & \underline{3319} & \textbf{1.11} \\
        \midrule
        \multicolumn{22}{c}{\textbf{7B Model}} \\
        \midrule
        DS-7B & 87.5 & 676 & 0.00 & 90.8 & 3713 & 0.00 & 85.0 & 5963 & 0.00 & 64.4 & 7363 & 0.00 & 51.5 & 10485 & 0.00 & 37.1 & 11140 & 0.00 & 69.4 & 6557 & 0.00 \\
        $\heartsuit$ AutoThink & 91.1 & 866 & -0.16 & 91.2 & 2146 & 0.44 & 83.3 & 4645 & 0.12 & 65.5 & 5133 & 0.35 & 54.8 & 8051 & 0.42 & 36.2 & 8608 & 0.11 & 70.4 & 4908 & 0.21 \\
        $\heartsuit$ AdaptThink & 91.0 & \underline{309} & \underline{0.66} & 92.0 & 1875 & 0.54 & 87.5 & 3287 & 0.54 & 65.5 & 5574 & 0.29 & 55.6 & 8599 & 0.42 & 36.2 & 9523 & 0.03 & 71.3 & 4861 & 0.41 \\
        $\heartsuit$ DEER$^{\ddagger}$ & 90.6 & 917 & -0.25 & 89.8 & 2143 & 0.37 & 85.0 & 4451 & 0.25 & 52.6 & 5420 & -0.65 & 49.2 & 9839 & -0.16 & 36.7 & 7257 & 0.30 & 67.3 & 5005 & -0.02 \\
        $\spadesuit$ DIET & -- & -- & -- & 92.1 & 3187 & 0.19 & 82.6 & 6075 & -0.16 & -- & -- & -- & 57.9 & 10124 & 0.41 & -- & -- & -- & -- & -- & -- \\
        $\spadesuit$ ACPO & 88.3 & 413 & 0.42 & 91.6 & 1405 & 0.65 & -- & -- & -- & -- & -- & -- & 52.8 & \underline{4520} & 0.64 & -- & -- & -- & -- & -- & -- \\
        $\spadesuit$ Laser-D & 87.3 & 804 & -0.20 & 92.2 & 1836 & 0.55 & 90.0 & 2694 & 0.72 & 66.7 & 3914 & 0.58 & 58.3 & 5379 & \textbf{0.88} & 38.0 & 6167 & 0.52 & 72.1 & 3466 & 0.51 \\
        $\spadesuit$ Laser-DE & 82.2 & 789 & -0.47 & 92.0 & 1658 & 0.59 & 89.1 & 2612 & 0.71 & 66.8 & 3643 & 0.62 & 55.8 & 4969 & 0.78 & 37.5 & 6077 & 0.49 & 70.6 & 3291 & 0.45 \\
        $\spadesuit$ SmartThinker & -- & -- & -- & 93.0 & 2753 & 0.33 & 90.0 & 4118 & 0.49 & -- & -- & -- & -- & -- & -- & 40.8 & 9277 & 0.47 & -- & -- & -- \\
        $\spadesuit$ Step Entropy$^{\dagger}$ & 79.2 & 168 & 0.28 & 85.0 & 2439 & 0.03 & -- & -- & -- & -- & -- & -- & 57.1 & 6839 & 0.68 & 33.3 & 10750 & -0.47 & -- & -- & -- \\
        $\clubsuit$ LC-R1 & 88.1 & 450 & 0.36 & 90.4 & 1568 & 0.56 & 79.1 & 3453 & 0.07 & 64.1 & 4144 & 0.41 & 23.6 & 6904 & -2.37 & 36.2 & 7150 & 0.24 & 63.6 & 3945 & -0.12 \\
        $\clubsuit$ S-GRPO & \textbf{93.8} & 906 & -0.12 & 92.4 & 2252 & 0.45 & 87.5 & 3494 & 0.50 & 69.7 & 3914 & 0.71 & 56.0 & 7377 & 0.56 & 36.0 & 7908 & 0.14 & 72.6 & 4308 & 0.37 \\
        $\clubsuit$ VSRM-R++ & -- & -- & -- & 89.8 & 2044 & 0.40 & 80.9 & 3704 & 0.14 & 66.1 & 5470 & 0.34 & 52.2 & 6773 & 0.39 & 36.4 & 6953 & 0.28 & -- & -- & -- \\
        $\clubsuit$ DEPO & -- & -- & -- & \textbf{94.4} & 2318 & 0.50 & 90.5 & 3215 & 0.65 & -- & -- & -- & 52.7 & 6580 & 0.44 & 39.2 & 7092 & 0.53 & -- & -- & -- \\
        $\clubsuit$ DECS & -- & -- & -- & 93.0 & 1728 & 0.61 & 89.0 & 2772 & 0.68 & \underline{70.3} & 3283 & \underline{0.83} & 51.3 & 5277 & 0.48 & 36.4 & 5516 & 0.41 & -- & -- & -- \\
        $\clubsuit$ SWAP & -- & -- & -- & 93.6 & 1573 & 0.67 & 85.6 & 2685 & 0.57 & 60.3 & 4760 & 0.03 & 56.5 & 5744 & 0.74 & 38.1 & 5598 & 0.58 & -- & -- & -- \\
        $\clubsuit$ SPAE & -- & -- & -- & -- & -- & -- & \textbf{92.5} & 4543 & 0.50 & \textbf{74.0} & 6509 & 0.56 & \textbf{59.4} & 9908 & 0.51 & \textbf{42.7} & 10687 & 0.50 & -- & -- & -- \\
        $\spadesuit$ TLMRE & 91.1 & 515 & 0.36 & \underline{93.8} & 1700 & 0.64 & 89.9 & 2967 & 0.68 & 69.6 & 4454 & 0.64 & \underline{58.5} & 7462 & 0.70 & 39.2 & 8301 & 0.43 & \underline{73.7} & 4233 & 0.57 \\
        $\clubsuit$ AttnPO & 92.4 & 446 & 0.51 & 92.8 & \underline{1300} & \underline{0.72} & 89.6 & \underline{2303} & \underline{0.78} & 68.7 & \underline{2912} & 0.80 & 57.2 & 5214 & 0.83 & 38.1 & \underline{5359} & \underline{0.60} & 73.1 & \underline{2922} & \underline{0.71} \\
        \midrule
        $\clubsuit$ \textbf{Ours} & \underline{92.9} & \textbf{264} & \textbf{0.79} & 91.9 & \textbf{948} & \textbf{0.78} & \underline{91.0} & \textbf{1731} & \textbf{0.92} & 68.5 & \textbf{2370} & \textbf{0.87} & 56.3 & \textbf{4356} & \underline{0.86} & \underline{42.6} & \textbf{5275} & \textbf{0.97} & \textbf{73.9} & \textbf{2491} & \textbf{0.87} \\
        \bottomrule
    \end{tabular*}
    \endgroup
    \caption{Acc., Tok. and AES across six mathematical reasoning benchmarks. \textbf{Bold} indicate the best results and \underline{underlined} indicate the second-best results. $\heartsuit$, $\spadesuit$, and $\clubsuit$ denote adaptive reasoning, outcome-supervised RL, and process-supervised RL methods, respectively. $^{\dagger}$Step Entropy counts thinking tokens only, so its Tok. and AES are excluded from highlighting. $^{\ddagger}$DEER uses greedy decoding.}
    \label{tab:main-results}
\end{table*}
\begin{table}[!t]
    \centering
    \fontsize{7}{9}\selectfont
    \setlength{\tabcolsep}{1.5pt}
    \renewcommand{\arraystretch}{1.0}
    \begin{tabular*}{\columnwidth}{@{\extracolsep{\fill}}l*{9}{c}@{}}
        \toprule
        \multirow{2}{*}{\textbf{Model}} & \multicolumn{3}{c}{\textbf{LiveCodeBench}} & \multicolumn{3}{c}{\textbf{GPQA}} & \multicolumn{3}{c}{\textbf{MMLU}} \\
        \cmidrule(l{1.5pt}r{1.5pt}){2-4} \cmidrule(l{1.5pt}r{1.5pt}){5-7} \cmidrule(l{1.5pt}r{1.5pt}){8-10}
        & Acc. & Tok. & AES & Acc. & Tok. & AES & Acc. & Tok. & AES \\
        \midrule
        \multicolumn{10}{c}{\textbf{1.5B Model}} \\
        \midrule
        DS-1.5B & 24.0 & 10286 & 0.00 & 27.9 & 9603 & 0.00 & 52.1 & 2650 & 0.00 \\
        \textbf{Ours} & \textbf{28.6} & \textbf{6656} & \textbf{0.93} & \textbf{33.1} & \textbf{4889} & \textbf{1.05} & \textbf{52.9} & \textbf{1357} & \textbf{0.53} \\
        \midrule
        \multicolumn{10}{c}{\textbf{7B Model}} \\
        \midrule
        DS-7B & 48.1 & 8905 & 0.00 & \textbf{49.2} & 8153 & 0.00 & 71.6 & 2067 & 0.00 \\
        \textbf{Ours} & \textbf{48.5} & \textbf{5436} & \textbf{0.41} & 47.9 & \textbf{4029} & \textbf{0.36} & \textbf{71.7} & \textbf{803} & \textbf{0.62} \\
        \bottomrule
    \end{tabular*}
    \caption{OOD results on LiveCodeBench v6, GPQA Diamond, and MMLU. Best results are in \textbf{bold}.}
    \label{tab:ood-results}
\end{table}

\subsection{Effectiveness of IGPO}
\label{sec:main-results}
\paragraph{Main Results.}
Table~\ref{tab:main-results} compares IGPO with existing efficient reasoning methods across six mathematical reasoning benchmarks.
IGPO achieves the highest macro-average accuracy and AES among the compared methods at both model scales.
Relative to the base models, it reduces average response length by up to 62\% while improving accuracy.
Compared with AttnPO and TLMRE, IGPO achieves a better overall accuracy–efficiency trade-off on DS-1.5B. 
Its advantage is more pronounced on DS-7B, where it improves macro-average accuracy while substantially reducing average response length, producing shorter responses than both baselines on all six benchmarks. 
In addition, on the challenging AIME2025 benchmark, where existing baselines struggle to balance accuracy and reasoning cost, IGPO substantially improves this trade-off on DS-7B, achieving an AES of 0.97 versus 0.60 for the best baseline.
Overall, these results suggest that our approach achieves a favorable trade-off between reasoning efficiency and task performance.

\paragraph{OOD Generalization.}
Table~\ref{tab:ood-results} shows that IGPO's efficiency gains extend beyond mathematical reasoning, reducing response length by 35.3--61.2\% across all three OOD benchmarks and both model scales. These savings accompany improved performance on DS-1.5B and largely preserved performance on DS-7B, with a 1.3~pp decrease on GPQA.\footnote{Note that IGPO is trained exclusively on mathematical reasoning data, so limited cross-domain gains are not unexpected. We will explore multi-domain training in the future.} 
All settings achieve positive AES, indicating favorable accuracy--efficiency trade-offs beyond the training domain.

\begin{figure*}[!t]
    \centering
    \includegraphics[width=\textwidth]{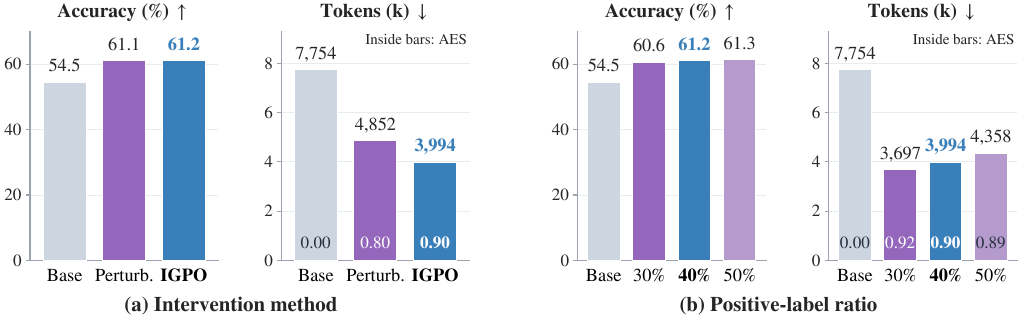}
    \caption{
        Ablation and sensitivity analysis on DS-1.5B.
        \textbf{(a)} Intervention-method ablation:
        simple perturbation (Perturb.) versus unlearning (IGPO).
        \textbf{(b)} Sensitivity to the positive-label ratio $\rho$.
        Values inside token bars denote AES.
        Blue bars indicate IGPO with the default $\rho=40\%$.
    }
    \label{fig:ablation-sensitivity}
\end{figure*}

\begin{figure}[t]
    \centering
    \includegraphics[width=\columnwidth]{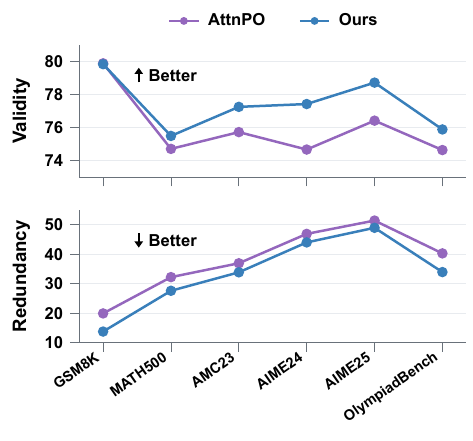}
    \caption{Average validity and redundancy scores ($\times 100$) for 7B models, evaluated with ReasonEval-7B.}
    \label{fig:reasoneval}
\end{figure}

\subsection{CoT Quality Evaluation}
\label{sec:reasoning-quality-analysis}
To assess the quality of IGPO's shortened CoT, we use ReasonEval-7B~\citep{xia2025reasoneval} to evaluate CoT generated by IGPO and AttnPO on DS-7B across the six mathematical reasoning benchmarks. 
Following its scoring protocol, we report average \textit{validity} and \textit{redundancy} scores.
Validity assesses the computational correctness and logical soundness of CoT steps. 
As shown in Figure~\ref{fig:reasoneval}, IGPO achieves higher validity than AttnPO on most benchmarks, with particularly clear gains on AIME2024 and AIME2025.
Redundancy measures correct steps that do not advance the solution. 
IGPO achieves lower redundancy than AttnPO on all datasets. 
These results suggest that IGPO produces shorter, less redundant CoT while maintaining or improving step validity.

\subsection{Ablation and Sensitivity Analysis}
\label{sec:ablations}
\label{sec:sensitivity-analysis}
For computational efficiency, we conduct these analyses on DS-1.5B with a batch size of 64 and 8 rollouts per question, using the same configuration across variants. Unless otherwise specified, other hyperparameters match the main experiments.

\paragraph{Intervention Method.}
We replace step-targeted unlearning with simple perturbation following \citet{zhao2026ahamoments}, while keeping the fraction of steps labeled important unchanged. Figure~\ref{fig:ablation-sensitivity}(a) shows that unlearning achieves similar accuracy but shorter responses, increasing AES from 0.80 to 0.90. 
These results suggest that unlearning-derived credit provides a more favorable accuracy--efficiency trade-off under the same training configuration.

\paragraph{Impact of Positive-Label Ratio.}
We vary the global high-ISC ratio $\rho$ over $\{30\%, 40\%, 50\%\}$ and perform RL accordingly.
As shown in Figure~\ref{fig:ablation-sensitivity}(b), larger $\rho$ yields higher accuracy but longer responses, whereas smaller $\rho$ favors stronger compression. 
The default $\rho = 40\%$ provides an intermediate operating point between two objectives.

\paragraph{Case Study.} Appendix~\ref{app:case-studies} compares reasoning traces from IGPO and AttnPO. 
IGPO avoids repeated geometric reasoning and self-verification in the first example, and computes the required product directly in the second without unnecessary detours or post-answer verification, using approximately 39--60\% fewer CoT tokens while retaining the key derivations. These examples suggest that IGPO more reliably identifies redundant reasoning than AttnPO, leading to more concise responses.

\section{Conclusion}
\label{sec:conclusion}
We introduced ISC and IGPO for learning efficient reasoning from intervention-derived step credit. ISC assesses step influence through targeted parameter interventions under fixed reasoning prefixes, while IGPO distills this credit to guide stepwise advantage rescaling. Experiments across two model scales and nine benchmarks demonstrate favorable accuracy--efficiency trade-offs, with up to 62\% token reduction on mathematical reasoning tasks.
Further analysis suggests that IGPO generates more concise CoT traces with more reliable computations and logical deductions.
Future work will explore reducing offline annotation costs and extending training to broader domains.

\section*{Limitations}
This work has the following limitations: 1) Step-level parameter unlearning approximates interventions on internal computation, so interpreting ISC depends on the effectiveness and selectivity of unlearning; 2) Estimating ISC requires independent parameter interventions for individual steps, incurring a one-time offline annotation cost. Distillation avoids online interventions but does not eliminate this cost; 3) Our training focuses on mathematical reasoning with verifiable answers. Extending the framework to more domains and open-ended tasks without clear verification criteria requires further research.

\appendix
\section{Supplementary Method Details}
\label{app:details}
\subsection{Prompts}
\label{app:prompts}
For rollout generation, we provide only the mathematical question.
For answer prediction during ISC annotation, we append the following suffix to $h_n$:
\begin{verbatim}
**Final Answer**
\boxed{
\end{verbatim}
We score the answer generated by the base model, averaging log-probabilities only over the content inside \verb|\boxed{...}|.

\subsection{Step Segmentation}
\label{app:step-segmentation}
We use double-newline (\verb|\n\n|) segmentation throughout annotation, predictor training, and policy optimization.
Each delimiter belongs to the preceding step.
The final answer-bearing step within the CoT and all text after \verb|</think>| are excluded from both unlearning and advantage rescaling.
These tokens retain the original trajectory-level advantage during RL.

\subsection{Unlearning Objective}
\label{app:annotation-details}
The NPO term~\citep{zhang2024npo} in Eq.~\eqref{eq:step-unlearning-objective} is
\begin{equation}
    \begin{aligned}
        &\mathcal{L}_{\mathrm{NPO}}^{(n)}(\theta) \\
        &=-\frac{2}{\beta}\log\sigma\!\left[
            \beta\log\frac{\pi_{\theta_0}(s_n\mid h_n)}
                     {\pi_{\theta}(s_n\mid h_n)}
        \right].
    \end{aligned}
    \label{eq:step-npo-loss}
\end{equation}
The step log-likelihood sums token log-probabilities without length normalization.
For retain regularization, each update samples one other correct trajectory.
Let $\mathcal{R}$ index its response tokens, including the CoT and final answer, and let $x_t$ denote the context preceding token $y_t$.
We use a token-averaged $k_3$ surrogate for KL regularization:
\begin{equation}
    \begin{aligned}
        d_t&=\log\frac{\pi_{\theta_0}(y_t\mid x_t)}
                         {\pi_{\theta}(y_t\mid x_t)}, \\
        \mathcal{L}_{\mathrm{KL}}(\theta)
        &=\frac{1}{|\mathcal{R}|}\sum_{t\in\mathcal{R}}
          \left(e^{d_t}-d_t-1\right).
    \end{aligned}
    \label{eq:retain-kl-loss}
\end{equation}
The implementation clips log-ratios and penalties for numerical stability.
This regularizer is evaluated on fixed retain trajectories and combined with NPO in Eq.~\eqref{eq:step-unlearning-objective}, with $\lambda_{\mathrm{KL}}=1$.

\subsection{Predictor Objective}
\label{app:predictor-details}
The shared linear classification head reads the final-layer hidden state at the last token of each step.
For a batch $\mathcal{B}$, let $\mathcal{V}_i$ denote the valid annotated steps in trajectory $i$.
The training objective is
\begin{equation}
    \mathcal{L}_{\mathrm{pred}}(\phi)
    =\frac{1}{|\mathcal{B}|}\sum_{i\in\mathcal{B}}
    \frac{1}{|\mathcal{V}_i|}\sum_{n\in\mathcal{V}_i}
    \operatorname{BCE}(p_{i,n},z_{i,n}).
    \label{eq:ISC-predictor-loss}
\end{equation}
Thus, each trajectory receives equal total weight.

\subsection{Policy Objective}
\label{app:policy-details}
We optimize the RLOO advantages~\citep{ahmadian2024back} after ISC rescaling using a PPO clipped policy loss~\citep{schulman2017ppo}.
Let $w_{i,t}(\psi)=\pi_{\psi}(o_{i,t}\mid q,o_{i,<t})/\pi_{\psi_{\mathrm{old}}}(o_{i,t}\mid q,o_{i,<t})$ denote the probability ratio between the current and rollout policies.
The token-level loss before dual clipping is
\begin{equation}
    \begin{aligned}
        \ell_{i,t}=-\min\Bigl(&w_{i,t}(\psi)\widetilde{A}_{i,t}, \\
        &\operatorname{clip}\bigl(w_{i,t}(\psi),1-\epsilon,1+\epsilon\bigr)
        \widetilde{A}_{i,t}\Bigr),
    \end{aligned}
    \label{eq:policy-clipped-surrogate}
\end{equation}
where $\epsilon=0.2$.
For negative advantages, dual clipping caps this loss at $-c\widetilde{A}_{i,t}$ with $c=3.0$.

\section{Experimental Settings}
\label{app:experimental-settings}
\subsection{Training Hyperparameters}
\label{app:training-details}
Table~\ref{tab:training-configuration} summarizes the hyperparameters for ISC annotation, predictor training, and policy optimization.

\begin{table*}[!t]
    \centering
    \begin{tabularx}{\textwidth}{@{}lXX@{}}
        \toprule
        Stage & Setting & Value \\
        \midrule
        ISC annotation & Updated modules & LoRA~\citep{hu2022lora} in \texttt{down\_proj} \\
        & High-ISC proportion $\rho$ & Global $0.4$ \\
        & LoRA rank & $32$ \\
        & Learning rate and update count & $5\times10^{-4}$, 5 updates per target step \\
        & Optimizer & AdamW \\
        & NPO $\beta$ and retain weight $\lambda_{\mathrm{KL}}$ & $\beta=0.1$, $\lambda_{\mathrm{KL}}=1.0$ \\
        & Retain input length limit & 8,192 tokens \\
        & Rollout temperature & $0.6$ (DS-1.5B), $1.0$ (DS-7B) \\
        & Maximum rollout length & 16,384 tokens \\
        \midrule
        ISC predictor & Backbone & Qwen3-4B-Instruct-2507 \\
        & Trainable parameters & Entire backbone and classification head \\
        & Train/validation split & 900/100 trajectories per model scale \\
        & Training epochs & $10$ \\
        & Prediction threshold $\eta$ & $0.5$ \\
        & Optimizer & AdamW \\
        & Learning rate and batch size & $2\times10^{-5}$, $32$ \\
        & Maximum sequence length & 17,408 tokens \\
        \midrule
        Policy optimization & Advantage scaling $\delta$ & $0.7$ (DS-1.5B), $0.5$ (DS-7B) \\
        & Length weight $\alpha$ & $0.15$ for both model scales \\
        & Rollout group size $G$ & $16$ \\
        & PPO clipping threshold $\epsilon$ / dual-clip $c$ & $0.2$ / $3.0$ \\
        & KL regularization / entropy bonus & Disabled / disabled \\
        & Optimizer & AdamW \\
        & Learning rate & $2\times10^{-6}$ \\
        & Training steps & 1,200 \\
        & Batch size & $128$ \\
        & Maximum response length & 16,384 tokens \\
        & Rollout temperature & $1.0$ \\
        \bottomrule
    \end{tabularx}
    \caption{Training hyperparameters.}
    \label{tab:training-configuration}
\end{table*}

\subsection{Evaluation Settings and AES}
\label{app:evaluation-protocol}
For mathematical reasoning evaluation, we sample 128 responses per question on AIME2024 and AIME2025 and four on the other benchmarks.
We use a sampling temperature of $0.6$, top-$p=0.95$, and a maximum generation length of 16,384 tokens.

Let $A$ and $L$ denote accuracy and average response length, with $A_b$ and $L_b$ the corresponding base-model values.
We compute AES~\citep{luo2025o1pruner} as
\begin{equation}
    \operatorname{AES}=1-\frac{L}{L_b}+
    \begin{cases}
        3(A/A_b-1), & A\geq A_b, \\
        5(A/A_b-1), & A<A_b.
    \end{cases}
    \label{eq:aes}
\end{equation}
Macro averages assign equal weight to each benchmark, with AES computed separately for each benchmark before averaging.

\section{Case Studies}
\label{app:case-studies}
Figures~\ref{fig:case-math500} and~\ref{fig:case-aime2024} compare the 7B models on MATH500 and AIME2024.
Red marks redundant passages, and green highlights derived results from which the model can advance toward the answer.
These highlights are manually annotated.

\definecolor{caseRed}{RGB}{214,39,40}
\definecolor{caseGreen}{RGB}{0,137,53}

\begin{figure*}[!t]
    \centering
    \begingroup
    \setlength{\fboxsep}{8pt}
    \colorbox{black!5}{%
        \begin{minipage}{\dimexpr\textwidth-2\fboxsep\relax}
            \small
            \setlength{\parindent}{0pt}
            \setlength{\parskip}{4pt}
            \setlength{\fboxsep}{3pt}
            \textbf{Question (MATH500).}
A steel sphere with a 3-inch radius is made by removing metal from the corners of a cube that has the shortest possible side lengths. How many cubic inches are in the volume of the cube?

\medskip
\textbf{AttnPO-7B (401 CoT tokens).}

Okay, so I have this problem here: A steel sphere with a 3-inch radius is made by removing metal from the corners of a cube that has the shortest possible side lengths. How many cubic inches are in the volume of the cube?

Hmm, let me try to visualize this. There's a cube, and from each corner, metal is removed to shape a sphere inside. The sphere has a radius of 3 inches, so its diameter is 6 inches. I think the sphere is inscribed inside the cube, touching the centers of each face of the cube. If that's the case, then the diameter of the sphere should be equal to the side length of the cube.

\textcolor{caseRed}{Wait, but the problem says the sphere is made by removing metal from the corners of the cube. So maybe the sphere is inscribed such that it touches the centers of each face? That would mean the diameter of the sphere equals the side length of the cube.}

But the sphere has a radius of 3 inches, so diameter is 6 inches. Therefore, the cube must have a side length of 6 inches. \textcolor{caseGreen}{Then the volume would be $6^3 = 216$ cubic inches.}

\textcolor{caseRed}{But hold on, the question says ``the shortest possible side lengths''. So maybe there's a different configuration? If you remove metal from the corners of a cube to make a sphere, perhaps the sphere is circumscribed around the cube? No, that doesn't make sense because the sphere would then have a larger diameter.}

\textcolor{caseRed}{Wait, maybe I need to think differently. If the sphere is formed by removing metal from the cube's corners, the sphere must fit perfectly inside the cube, touching the centers of each face. So the sphere's diameter equals the cube's edge length. So radius 3 inches means diameter 6 inches, cube edge length 6 inches, volume 216.}

\textbf{Final Answer}\par
The volume of the cube is $\boxed{216}$ cubic inches.

\medskip
\textbf{IGPO-7B (162 CoT tokens).}

Okay, so I have this problem here: A steel sphere with a 3-inch radius is made by removing metal from the corners of a cube that has the shortest possible side lengths. How many cubic inches are in the volume of the cube?

Hmm, let's see. So, the sphere is made by removing metal from the corners of a cube. That probably means the sphere is inscribed inside the cube, touching the centers of each face of the cube. So the diameter of the sphere would be equal to the side length of the cube.

Given the radius is 3 inches, so diameter is 6 inches. Therefore, the cube has side length 6 inches. \textcolor{caseGreen}{So volume is $6^3 = 216$ cubic inches.}

\textbf{Final Answer}\par
$\boxed{216}$
        \end{minipage}%
    }
    \endgroup
    \caption{A MATH500 case: AttnPO exhibits repeated reasoning and unnecessary self-verification, whereas IGPO avoids these redundancies.}
    \label{fig:case-math500}
\end{figure*}

\begin{figure*}[!t]
    \centering
    \begingroup
    \setlength{\fboxsep}{8pt}
    \colorbox{black!5}{%
        \begin{minipage}{\dimexpr\textwidth-2\fboxsep\relax}
            \small
            \setlength{\parindent}{0pt}
            \setlength{\parskip}{4pt}
            \setlength{\fboxsep}{3pt}
            \textbf{Question (AIME2024).}
There exist real numbers $x$ and $y$, both greater than 1, such that $\log_x(y^x) = \log_y(x^{4y}) = 10$. Find $xy$.

\medskip
\textbf{AttnPO-7B (716 CoT tokens).}

Okay, so I have this problem here: There exist real numbers $x$ and $y$, both greater than 1, such that log base $x$ of $y^x$ is equal to 10, and log base $y$ of $x^{4y}$ is also equal to 10. I need to find the product $xy$. Hmm, let me write down what I know. First, the equation $\log_x(y^x) = 10$. I can rewrite this using logarithm properties. Remember that $\log_b(a^c) = c\cdot \log_b(a)$, so this becomes $x\cdot \log_x(y) = 10$. Similarly, the second equation $\log_y(x^{4y}) = 10$ can be rewritten as $4y\cdot \log_y(x) = 10$. So now I have two equations: 1) $x\cdot \log_x(y) = 10$ 2) $4y\cdot \log_y(x) = 10$ Hmm, maybe I can express these in terms of natural logarithms or something? Or perhaps relate them using change of base formula. I know that $\log_b(a) = 1 / \log_a(b)$. So $\log_x(y) = 1 / \log_y(x)$. Let me denote $\log_y(x)$ as $k$. Then $\log_x(y)$ is $1/k$. So substituting into the first equation: $x\cdot (1/k) = 10 \Rightarrow x = 10k$. And the second equation: $4y\cdot k = 10 \Rightarrow y\cdot k = 10/4 = 2.5 \Rightarrow y = 2.5 / k$. \textcolor{caseGreen}{So now I have $x = 10k$ and $y = 2.5 / k$.}

\textcolor{caseRed}{I need another equation to relate $x$ and $y$. Maybe I can use the definition of logarithms or another relation. Wait, from the change of base formula, $\log_y(x) = k$, which means $x = y^k$. But I also have $x = 10k$ and $y = 2.5 / k$. So substituting $x = 10k$ into $x = y^k$: $10k = (2.5 / k)^k$. Hmm, this seems complicated. Maybe another approach.}

Alternatively, since $x = 10k$ and $y = 2.5/k$, I can multiply $x$ and $y$: \textcolor{caseGreen}{$xy = (10k)\cdot (2.5 / k) = 10\cdot 2.5 = 25$.}

\textcolor{caseRed}{Wait, is that possible? The $k$ cancels out, so $xy = 25$ regardless of $k$? That seems too straightforward. But let me verify. If $xy = 25$, then maybe I can check if this satisfies the original equations. Suppose $xy = 25$, so $y = 25/x$. Plug into the first equation: $\log_x(y^x) = 10$. Which is $x\cdot \log_x(y) = 10$. Since $y = 25/x$, $\log_x(25/x) = \log_x(25) - \log_x(x) = \log_x(25) - 1$. So $x\cdot (\log_x(25) - 1) = 10$. Hmm, not sure if this helps. Wait, maybe my initial approach was correct. Since $x = 10k$ and $y = 2.5/k$, then $xy = 25$. So maybe the answer is 25.}

\textbf{Final Answer}\par
$\boxed{25}$

\medskip
\textbf{IGPO-7B (439 CoT tokens).}

Okay, so I need to solve this problem where there are real numbers $x$ and $y$, both greater than 1, such that $\log_x(y^x) = 10$ and $\log_y(x^{4y}) = 10$. I need to find $xy$. Alright, let's start by writing down the given equations: 1. $\log_x(y^x) = 10$ 2. $\log_y(x^{4y}) = 10$ I need to find $xy$. First, let's simplify both equations using logarithm properties. For the first equation: $\log_x(y^x) = 10$ Using the power rule: $x \cdot  \log_x(y) = 10$ So, $x \cdot  \log_x(y) = 10 \rightarrow$ Equation (1) Similarly, the second equation: $\log_y(x^{4y}) = 10$ Again, using power rule: $4y \cdot  \log_y(x) = 10 \rightarrow$ Equation (2) So now we have: (1) $x \cdot  \log_x(y) = 10$ (2) $4y \cdot  \log_y(x) = 10$ Let me recall that $\log_b(a) = 1 / \log_a(b)$. So $\log_x(y) = 1 / \log_y(x)$ Let's denote $\log_y(x)$ as $k$. Then $\log_x(y) = 1/k$ So substituting into Equation (1): $x \cdot  (1/k) = 10 \rightarrow x = 10k$ Equation (2): $4y \cdot  k = 10 \rightarrow 4y \cdot  k = 10 \rightarrow y = 10 / (4k) = 5 / (2k)$ \textcolor{caseGreen}{So now we have $x = 10k$ and $y = 5 / (2k)$} We need to find $xy$. \textcolor{caseGreen}{$xy = x \cdot  y = (10k) \cdot  (5 / (2k)) = (10k \cdot  5) / (2k) = (50k) / (2k) = 25$}

\textbf{Final Answer}\par
$\boxed{25}$
        \end{minipage}%
    }
    \endgroup
    \caption{An AIME2024 case: AttnPO takes an unnecessary reasoning detour and continues reflecting after deriving the answer, whereas IGPO avoids these behaviors.}
    \label{fig:case-aime2024}
\end{figure*}

\end{document}